\documentclass[10pt,twocolumn,letterpaper]{article}

\usepackage{iccv}
\usepackage{times}
\usepackage{epsfig}
\usepackage{graphicx}
\usepackage{amsmath}
\usepackage{amssymb}
\usepackage{color}
\usepackage{caption}
\usepackage{subcaption}
\usepackage{algpseudocode}
\usepackage{algorithm}
\usepackage{array}
\usepackage[numbers,sort]{natbib}
\usepackage{datetime}

\usepackage{tabularx}
\usepackage{lipsum}
\usepackage{pbox}

\usepackage[pagebackref=true,breaklinks=true,letterpaper=true,colorlinks,bookmarks=false]{hyperref}

\iccvfinalcopy 

\def\iccvPaperID{1499} 

\ificcvfinal\fi
\begin{document}

\title{Discovering Characteristic Landmarks on Ancient Coins\\using Convolutional Networks}

\author{Jongpil Kim and Vladimir Pavlovic\\
Rutgers, The State University of New Jersey\\
{\tt\small \{jpkim, vladimir\}@cs.rutgers.edu}
}

\maketitle

\begin{abstract}
In this paper, we propose a novel method to find characteristic landmarks on ancient Roman imperial coins using deep convolutional neural network models (CNNs).
We formulate an optimization problem to discover class-specific regions while guaranteeing specific controlled loss of accuracy.
Analysis on visualization of the discovered region confirms that not only can the proposed method successfully find a set of characteristic regions per class, , but also the discovered region is consistent with human expert annotations.
We also propose a new framework to recognize the Roman coins which exploits hierarchical structure of the ancient Roman coins using the state-of-the-art classification power of the CNNs adopted to a new task of coin classification.
Experimental results show that the proposed framework is able to effectively recognize the ancient Roman coins.
For this research, we have collected a new Roman coin dataset where all coins are annotated and consist of observe (head) and reverse (tail) images.
\end{abstract}

\section{Introduction}
The ancient Roman coins have not only bullion values from precious materials such as gold and silver, but also they provide people with beautiful and historical arts of relief.
They were first introduced during the third century BC and continued to be minted well across Imperial times.
The major role of the Roman coins was to make an exchange of goods and services easy for the Roman commerce.
Another important role in which researchers in numismatics have been interested is to convey historical events or news of the Roman empire via images on the coins.
Especially, the Roman imperial coins were used to provide political propaganda  across the empire by engraving portraits of the Roman emperors or important achievement of the empire.
As the Roman imperial coins are closely connect to the historical events of the empire, they could serve as importance references to understand the history of the Roman empire.

\begin{figure}
	\centering
	\begin{subfigure}[b]{0.47\linewidth}
		\includegraphics[width=\textwidth]{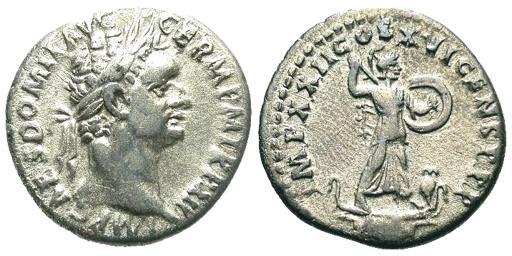}\\\centering
		observe~~~~~~~~~~~~reverse
		\caption{Domitian RIC 740}
	\end{subfigure}~~~~
	\begin{subfigure}[b]{0.47\linewidth}
		\includegraphics[width=\textwidth]{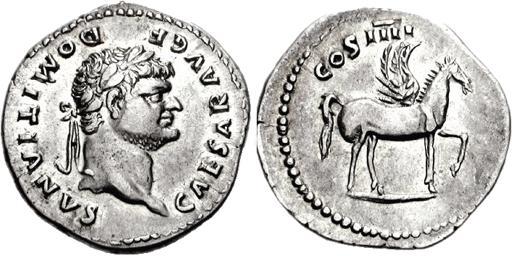}\\\centering
		observe~~~~~~~~~~~~reverse
		\caption{Domitian RIC 921}
	\end{subfigure}
	\caption{Sample observe and reverse images of two ancient Roman imperial coins. Both coins depict the same emperor (Domitian) on the observe side but have distinct reverse depictions, resulting in different Roman Imperial Coinage (RIC) labels. The descriptions for them are (a) Observe: \emph{Laureate head right}, Reverse: \emph{Minerva standing right on capital of rostral column with spear and shield to right owl}, and (b) Observe: \emph{Laureate head right}, Reverse: \emph{Pegasus right}.\label{fig:sample01}}
\end{figure}

In this paper, we aim at automatically finding visual characteristics of the ancient Roman imperial coins which make them distinguishable from the others, as well as recognizing their identities.
To achieve these goals, we collected Roman imperial coin images with their descriptions.
We used the Roman Imperial Coinage (RIC) \cite{Mattingly1923} to annotate the collected coin images.
RIC is a comprehensive numismatic catalog of Roman imperial currency which is the results of several decades of work.
The RIC provides a chronological catalog of the coins from 31 BC to 491 AD with description of both the obverse (head) and reverse (tail) sides of the coin.
Figure \ref{fig:sample01} shows example observe and reverse images and their descriptions.
For the purpose of the classification, we use the catalog number of RIC as a label to predict.

Automatic methods to identify the ancient coins have been attracted as a growing number of the coins are being traded everyday over the Internet \cite{Huber-Mork2011, Arandjelovic2010}.
One of the main issues in the active coin market is to prevent illegal trade and theft of the coins.
Traditionally, coin identification depends on manually searching catalogs of coin markets, auctions or the Internet.
However, it is impossible for the manual search to cover all trades because the coin market is very active, for example, over a half million coins are traded annually only in the north American market~\cite{Huber-Mork2011}.
Therefore, automatic identification of the ancient coins becomes significant.

Several works on coin classification have appeared in the computer vision field.
Some proposed methods use the edge detection of the engraved image on the coin~\cite{Maaten2006, Nolle2003}.
Others represent the coin images as local features such as SIFT~\cite{Lowe2004} and perform the classification~\cite{Kampel2008}.
Methods using the spatial pyramid models~\cite{Anwar2015} and orientations of pixels~\cite{Arandjelovic2010} are proposed to exploit the spatial information.
Aligning coin images using the deformable part models has refined the recognition accuracy over the standard spatial pyramid models~\cite{Kim2014}.

In this paper, we propose an automatic recognition method for the Roman imperial coins using the convolutional neural network models (CNNs).
Recently, the CNN models have shown the state-of-the-art performance in various computer vision problems including recognition, detection and segmentation \cite{Zeiler2014, Donahue2014, Zhang2014, Girshick2014}, driven by the increasing availability of large training dataset and the improvement of the computational power of the GPUs.
In this paper, we propose a hierarchical framework which employs the CNN models for the coin classification tasks by fine-tuning a pre-trained CNN model on the ImageNet dataset~\cite{Deng2009}.

Second, we propose a novel method to find characteristic landmarks on the coin images.
Our method is motivated by class saliency extraction proposed in \cite{Simonyan2014} to find class-sensitive regions.
In this paper, we formulate an optimization problem so that a minimal set of the parts on the coin image will be selected while the chosen set is still be recognized as the same category as the full image by the CNN models.
We consider the chosen parts are deemed the persistent, discriminative landmarks of the coin.
Such landmarks can be critical for analysis of coin features by domain experts, such as numismatists or historians.

The contributions of the paper can be highlighted as follows: 1) a new coin data set where all the coins have both observe (head) and reverse (tail) images with annotations, 2) a new framework of recognizing the Ancient Roman coins based on the CNNs, 3) a new optimization-based method to automatically find characteristic regions using the CNNs while guaranteeing specific controlled loss of accuracy.


\section{Related Work \label{s:rel}}

There have been several methods to recognize coins in the computer vision field.
Bag-of-words approaches with extracted visual descriptors for the coin recognition were proposed in \cite{Arandjelovic2010, Anwar2015, Anwar2014, Kim2014}.
A directional kernel to consider orientations of pixels~\cite{Arandjelovic2010} and an angle histogram method~\cite{Anwar2014} were proposed to use the explicit spatial information.
In \cite{Anwar2015}, rectangular spatial tiling, log-polar spatial tiling and circular spatial tiling methods were used to recognize the ancient coins.
Aligning the coin images by the deformable part model (DPM)~\cite{Felzenszwalb2010} further improves the recognition accuracy over the standard spatial pyramid model \cite{Kim2014}.
In this paper, we use the CNNs which exploit the spatial information by performing the convolution and handle the displacement of the coin image by performing the max-pooling.

The Roman imperial coin classification problem can be formulated as the fine-grained classification as all coins belong to one super class, {\it coin}.
To identify one class from the other looking-similar classes, which is one of the challenges in the fine-grained classification, people have conducted research on the part-based models so that objects are divided into a set of smaller parts and classification is performed by comparing the parts \cite{Gavves2013, Chai2013}.
However, those methods require annotated part labels while training, which takes an effort to obtain.
In this paper, we investigate an automatic method to find discriminative regions on the coins which does not depend on human's effort.


With the impressive performance of the deep convolutional neural network models, a lot of papers have been proposed to understand why and how they perform so well and give insight the behaviors of the internal layers.
The deconvolutional network~\cite{Zeiler2014} visualized the feature activities in the intermediate layers of the CNN models by mapping features to pixels in the reverse order.
A data-driven approach to visualize the receptive field of the neuron in the network was proposed in \cite{Zhou2015}.
The method in \cite{Zhou2015} is based on the exhaustive search using the sliding-window technique and measures the difference between presence and absence of one window on the coin image.
In \cite{Simonyan2014}, they propose an optimization method to reconstruct a representative image of a class from an empty image by calculating a gradient of the CNN model with respect to the image.
In this paper, we propose a novel method to find discriminative landmarks of the coin image by formulating on optimization problem.
Unlike \cite{Zhou2015} which requires exhaustive CNN evaluations for the sliding windows, our method effectively finds a set of discriminative regions by performing the optimization.

\section{Proposed Method \label{s:pro}}
In this section, we first describe how to train our convolutional neural network model for the task of the Roman imperial coin classification.
Then, we propose a novel method to discover characteristic landmarks which make one coin distinguishable from the others.

\subsection{Training Convolutional Neural Network for Coin Classification\label{s:pro1}}
The convolutional neural network (CNN) is the most popular deep learning method which was heavily studied in 1990s~\cite{Lecun1998}.
Recently, a large amount of labeled data and computational power using GPUs make it possible that the convolutional network becomes the most accurate object classification method~\cite{Krizhevsky2012}.

Let $S_c(\mathbf{\mathbf{x}})$ be the score of class $c$ for input $\mathbf{x}$, which is fed to a classification layer ({\it e.g.}, 1000-way softmax layer in \cite{Krizhevsky2012}).
Assuming that the softmax loss function is used, the loss function $\ell_c$ of the CNNs can be defined as:
\begin{equation}
\ell_c(\mathbf{x};\mathbf{w}) = -\log\left( \frac{\exp\left(S_{c}(\mathbf{x};\mathbf{w})\right)}{\sum_{c'} \exp\left(S_{c'}(\mathbf{x};\mathbf{w})\right) }\right), \label{eqn:cnn_loss}
\end{equation}
where $\mathbf{w}$ are the weights of the complex, highly structured, deep CNNs.
Then, stochastic gradient descent is used to minimize the loss $\ell_c$ by computing gradient with respect to $\mathbf{w}$ as $\partial \ell_c / \partial \mathbf{w}$.

Although the CNN models are successful when there exists large amount of labeled data, they are likely to perform poorly on small datasets because there are millions of parameters to be estimated~\cite{Zhang2013a}.
To overcome the limitation on the small data, a method to finetune the pre-trained model for new tasks was proposed, having shown successful performance~\cite{Long2015, Zhang2014, Zhou2014}.
In the fine-tunning method, we only need to change the softmax layer (which is the usually the last layer of the CNN models) appropriately for the new task.

Considering the number of the coin images in our dataset (about 4500), the CNN model is likely to be under-fitted if we train it only on the coin dataset even if we use the data augmentation method~\cite{Krizhevsky2012}.
Therefore, we train a deep convolution neural network (CNN) model in the fine-tuning manner.
To achieve the goal, we adopt one of the most popular architecture proposed by Krizhevsky et al. \cite{Krizhevsky2012} which is pre-trained on the ImagetNet with millions of natural images.
Specifically, we change the softmax layer of \cite{Krizhevsky2012} for our classification purpose, and then finetune the covolutional network under the supervised setting.
When training, we resize the original coin image to $256\times 256$ and randomly crop a sub region of size $224\times 224$ as the data augmentation discussed in  \cite{Krizhevsky2012}.
When testing, we crop the center of the coin.
We use the open-source package Caffe \cite{Jia2014} to implement our CNN model.

\subsection{Hierarchical Classification\label{s:pro_hi}}

\begin{figure}
	\begin{center}
		\includegraphics[width=0.60\linewidth]{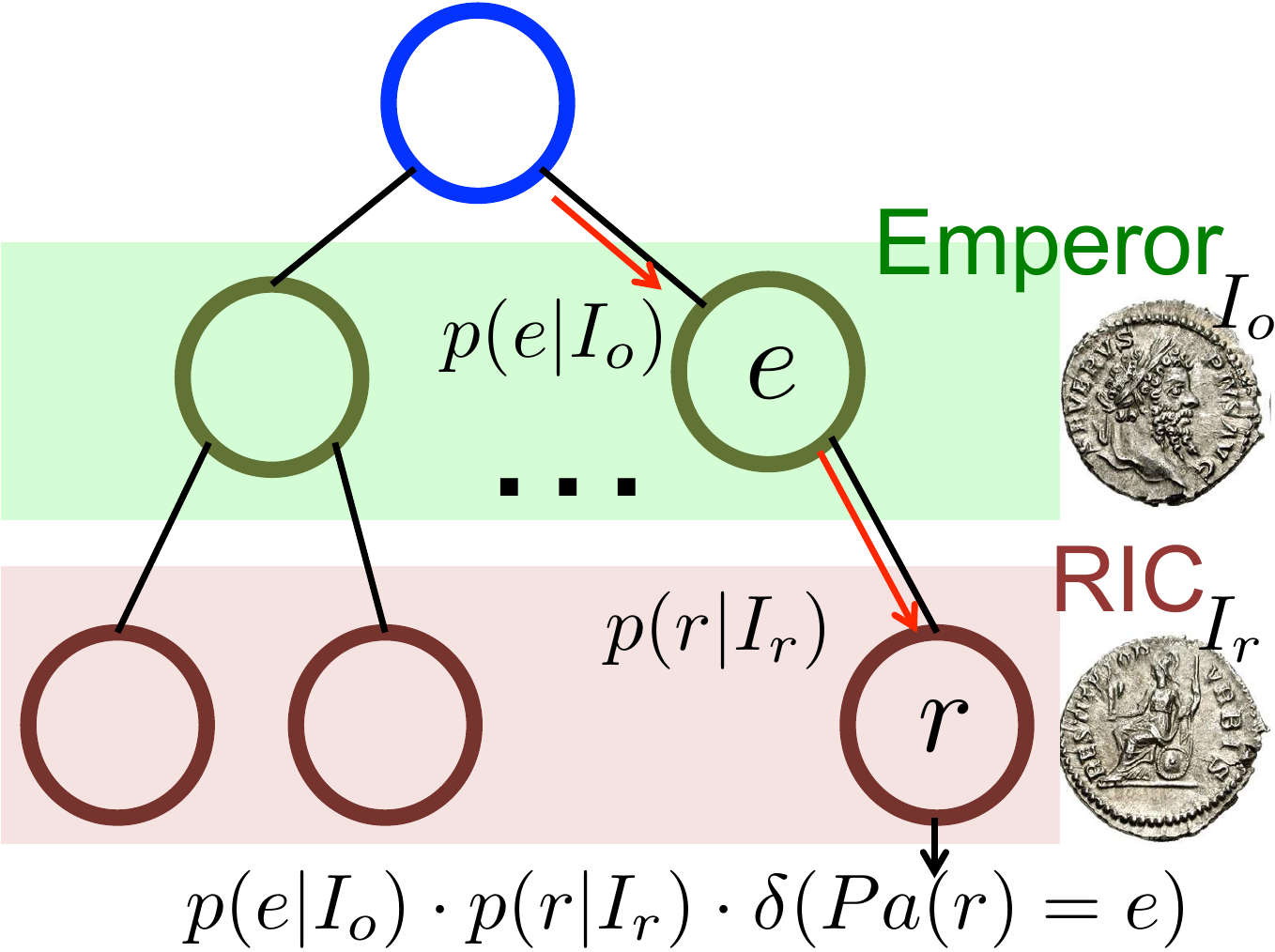}
		\caption{Hierarchical classification for the RIC label. $I_o$ and $I_r$ are the observe and reverse images, respectively. In the Emperor layer, we compute the probability of RIC label $r$ given reverse image $I_r$ (resp., in the RIC layer, probability of $e$ given $I_o$ ). Then the final prediction is defined as the product of the probabilities on the path from the root to the leaf. \label{fig:hier_class}}
	\end{center}
\end{figure}

Each coin in our dataset has both observe (head) and reverse (tail) images.
A straight-forward method to use both images is to feed them together to classifiers ({\it e.g.}, SVM or CNN) when training.
In this paper, we exploit a hierarchical structure of the Roman imperial coins.
One Roman emperor includes several RIC labels as shown in Figure \ref{fig:sample01} while one RIC label belongs to exactly one emperor.
Therefore, we can build a tree structure to represent the relationship between the Roman emperors and the RIC labels as depicted in Figure \ref{fig:hier_class}.
In the Emperor layer, we compute probability $p(e|I_o)$ for Emperor $e$ given observe image $I_o$, and in the RIC layer, $p(r|I_r)$ for RIC label $r$ given $I_r$.
Then the final probability is defined to be the product of the probabilities on the path from the root to the leaf as  $p(e|I_o) \cdot p(r|I_r) \cdot \delta(Pa(r) = e)$ where $Pa(r)$ is the parent of node $r$ and $\delta(\cdot)$ is the indicator function.

For this purpose, we train two CNN models, one for the RIC label taking the reverse image and the other for the Roman emperor taking observe image.
For a given pair observe and reverse images, we evaluate the probabilities on the nodes in the tree and choose the leaf node with the maximum value as the prediction result.

\subsection{Finding Characteristic Landmarks on Roman Coins \label{s:pro2}}
The coin classification problem can be considered as the fine-grained classification problem as all the images belong to one super class.
Finding discriminative regions that represent class characteristics plays an important role in the fine-grained classification.
This is specifically true in the context of Roman coins, where domain experts ({\it e.g.}, numismatists) seek intuitive, visual landmark feedback associated with an otherwise automated classification task.

In this section, we introduce our method to discover characteristic landmarks on the Roman coins using the CNN model.
We define the characteristic region set as {\it the smallest set of local patches} sufficient to represent the identity of the full image and distinguish it from other available classes.

Several approaches have been presented in the past that attempt to identify intuitive/visual class characteristics of CNNs \cite{Simonyan2014, Szegedy2014, Mahendran2014}.
However, their main purpose is  largely to reconstruct a representative, prototypical class image and not necessarily find the discriminative regions.
Unlike the previous methods, the proposed method starts from specific input image and removes visual information deemed irrelevant for the coin's accurate classification as an instance of the same class.

Let $\mathbf{I}$ and $\mathbf{I}(i)$ be the vectorized image and the $i$th pixel intensity of image $\mathbf{I}$, respectively.
Let $\mathbf{r}_k$, $1\leq k \leq K$, be the set of indices that belongs to the $k$th subregion in image $\mathbf{I}$.
The subregion could be a superpixel, a patch from the sliding window with overlapping, or even one pixel.
We define $\mathbf{I}_k$ to represent the $k$th subregion as follows:
\begin{equation}
\mathbf{I}_k(i) = \left\{\begin{array}{ll}\mathbf{I}(i)&\mbox{if $i \in \mathbf{r}_k$}\\0&\mbox{otherwise} \end{array}\right.,\qquad \bigcup_{k=1}^{K} \mathbf{I}_k = \mathbf{I}.
\end{equation}
Then we define a mask function $f_\mathbf{I}(\mathbf{x})$, $\mathbf{x} = [x_1, x_2, \ldots, x_k]^\top$ $x_i \in [0, 1]$, which maps image $\mathbf{I}$ to the masked image $f_\mathbf{I}(\mathbf{x})$ as a function of $\mathbf{x}$:
\begin{equation}
f_\mathbf{I}(\mathbf{x}) = \left(\sum_{k=1}^{K} x_k \cdot \mathbf{I}_k\right) \otimes \mathbf{C},
\end{equation}
where $\otimes$ is the element-wise product and $\mathbf{C}$ is a normalization vector counting how many times a pixel appears across the subregions as 
\begin{displaymath}
\mathbf{C}(i) = \frac{1}{\sum_{k=1}^{K} \delta(i \in \mathbf{r}_k)}.
\end{displaymath}
$x_k$ controls the transparency of the subregion $\mathbf{r}_k$ so that $x_k=1$ represents that the subregion has the full pixel intensity while $x_k=0$ implies that the region is transparent.

We would like to find an image such that the image consists of the smallest set of regions but still can be correctly classified by the original CNN model, with some small, controlled loss in confidence.
With the definition of $f_\mathbf{I}(\mathbf{x})$, we formulate our goal as follows:
\begin{align}
	\min_\mathbf{x} & \quad \ell_c\left(f_\mathbf{I} (\mathbf{x})\right) + \lambda  \mathcal{R}(\mathbf{x}) \label{eqn:opt02} \\
	\mbox{s.t.} &  \quad p\left(c|f_\mathbf{I}(\mathbf{x})\right) > p\left(c|f_\mathbf{I}(\mathbf{1})\right) - \epsilon, \label{eqn:opt03}\\
	& \quad \epsilon > 0, \nonumber
\end{align}
where $\ell_c(\cdot)$ is the loss function of the CNN model for class $c$, $\mathbf{1}$ is a vector of all ones, $\mathcal{R}(\cdot)$ is a regularization function and $\lambda$ is a hyper parameter to control the regularization.
We place the constraint so that the prediction probability of the masked image $f_\mathbf{I}(\mathbf{x})$ may differ from the original image $f_\mathbf{I}(\mathbf{1})$ at most $\epsilon$.

Because we are interested in absence or presence of a region,
the $L_0$-norm would be an ideal choice for the regularization function.
However, it is non-differentiable, making it difficult to optimize the objective function.
Therefore we resort to the $L_1$ norm which is the closest convex, $\mathcal{C}^0$ continuous approximation of the $L_0$-norm.

Both $\lambda$ and $\epsilon$ have similar roles to control the prediction accuracy of the masked image.
If we increase $\lambda$, $p\left(c|f_\mathbf{I}(\mathbf{\mathbf{x}})\right)$ decreases because the optimization puts more emphasis on minimizing the $|\mathbf{x}|_1$ than the loss function.
Similarly, large $\epsilon$ allows the low prediction accuracy of the masked image.
Therefore, in this paper we fix $\lambda$ to 1 and control $\epsilon$ because $\epsilon$ can explicitly put the lower bound of the prediction accuracy.

We use the negative log of the soft-max function as the loss:
\begin{displaymath}
\ell_c\left(f_{\mathbf{I}}(\mathbf{x})\right)  = -\log\left( \frac{\exp \left(S_c(f_{\mathbf{I}}(\mathbf{x})\right)}{\sum_{c'} \exp\left(S_{c'}(f_{\mathbf{I}}(\mathbf{x})\right)}\right),
\end{displaymath}
where $S_{c}$ is the score for class $c$ as in (\ref{eqn:cnn_loss}).

Optimization in (\ref{eqn:opt02}) is in general a non-convex problem, a consequence of the non-convex CNN mapping.
We approach the minimization task in (\ref{eqn:opt02}) using a general subgradient descent optimization with backprojection.
The gradient can be computed using the chain rule as:
\begin{eqnarray}
\frac{\partial \ell_c}{\partial \mathbf{x}} =\left(\frac{\partial f_{\mathbf{I}}(\mathbf{x})}{\partial \mathbf{x}}\right)^\top \cdot  \frac{\partial \ell_c}{\partial f_\mathbf{I}(\mathbf{x})} . \label{eqn:loss_wrt_x1}
\end{eqnarray}
The second component of the gradient, $ \partial \ell_c  / \partial f_{\mathbf{I}}(\mathbf{x})$, represents the sensitivity of the CNN output with respect to the input image (region) and can be computed by the back-propagation as discussed in \cite{Simonyan2014}.
Note that this quantity differs from the typical sensitivity of loss with respect to the CNN parameters, used in CNN training.
Because $f_{\mathbf{I}}(\mathbf{x})$ is a linear function of mask $\mathbf{x}$,
the gradient $\partial f_{\mathbf{I}}(\mathbf{x}) / \partial \mathbf{x}$ is easily computed as 
\begin{displaymath}
\begin{bmatrix}
\frac{\partial f_{\mathbf{I}}^1(\mathbf{x})}{\partial x_1} & \frac{\partial f_{\mathbf{I}}^1(\mathbf{x})}{\partial x_2} & \ldots &\frac{\partial f_{\mathbf{I}}^1(\mathbf{x})}{\partial x_K}\\
\frac{\partial f_{\mathbf{I}}^2(\mathbf{x})}{\partial x_1} & \frac{\partial f_{\mathbf{I}}^2(\mathbf{x})}{\partial x_2} & \ldots &\frac{\partial f_{\mathbf{I}}^2(\mathbf{x})}{\partial x_K}\\
\vdots & & \ddots& \vdots\\
\frac{\partial f_{\mathbf{I}}^K(\mathbf{x})}{\partial x_1} & \frac{\partial f_{\mathbf{I}}^K(\mathbf{x})}{\partial x_2} & \ldots &\frac{\partial f_{\mathbf{I}}^K(\mathbf{x})}{\partial x_K}
\end{bmatrix},
\end{displaymath}
where $f_{\mathbf{I}}^k(\mathbf{x})$ is the $k$th element of the masked image.

The standard gradient descent method to minimize (\ref{eqn:opt02}) may violate the constraint (\ref{eqn:opt03}) because of the regularization term that enforces sparseness.
Therefore, we use the backprojection method for the optimization.
We first initialize $\mathbf{x}$ to $\mathbf{1}$ ({\it i.e.}, full image), then perform the gradient descent.
If the violation occurs, we remedy it by taking the gradient with respect to only the loss function without considering the regularization until the constraint is satisfied.

During the optimization, the loss function and the $L_1$ regularization term in (\ref{eqn:opt02}) compete with each other under the constraint of (\ref{eqn:opt03}).
Minimization of the loss function alone typically requires a large number of regions.
On the other hand, the regularization term attempts to select as few landmarks as possible.
Because non-discriminative regions usually do not contribute to minimization of the loss function, they are more likely to be removed than the persistent, discriminative regions.

\section{Experiments and Results \label{s:exp}}
In this section, we explain our experimental settings including the coin data collection.
We then discuss the coin classification using the CNN model.
Finally, we analyze the results of discovering characteristic landmarks on the coin images.

\begin{figure}[t]
	\begin{center}
		\includegraphics[width=0.9\linewidth]{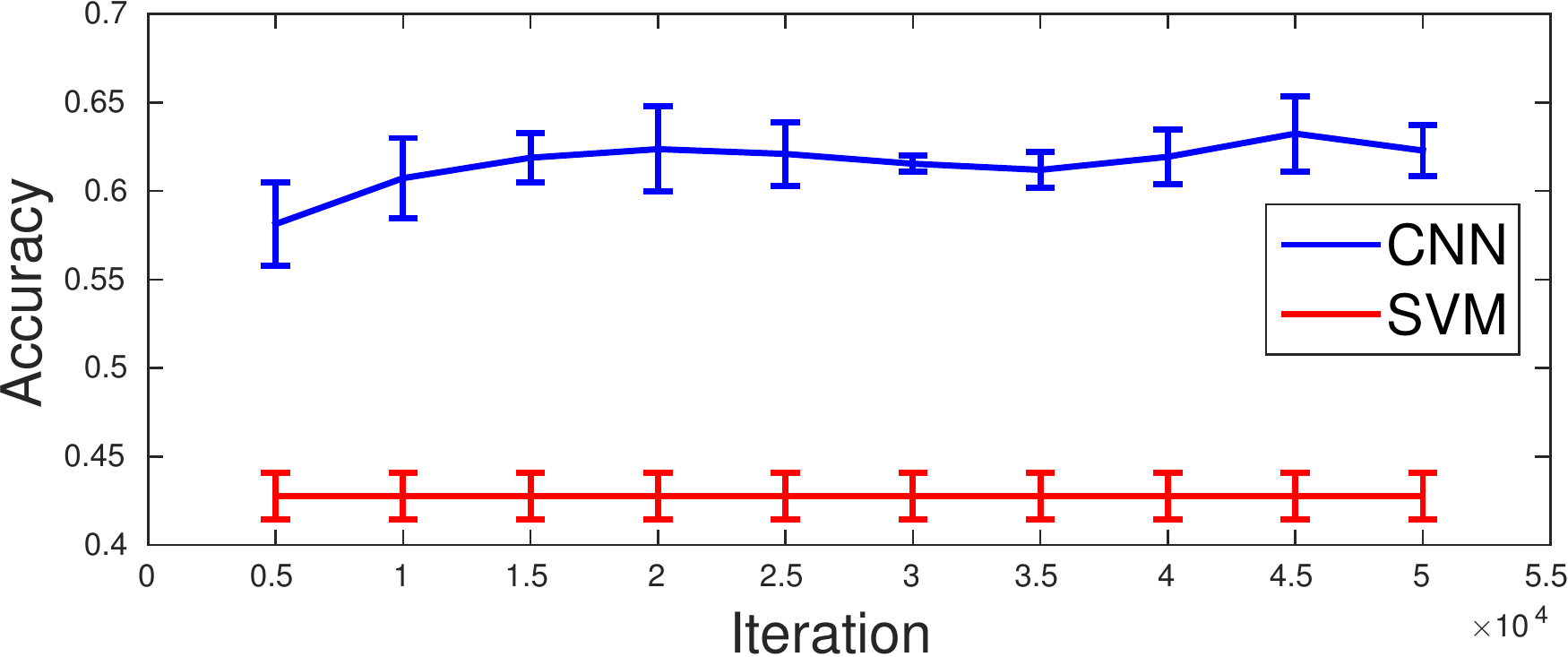}
		\caption{The change of the classification accuracy for \text{Reverse} as a function of the iteration number (epoch). After 40,000 iterations, the accuracy remains steady. Note that a small number of iterations is sufficient to outperform SVM.\label{fig:cnn_iter}}
	\end{center}
\end{figure}

\subsection{Experimental Settings}
{\bf Data collection}: 
We have collected ancient Roman Imperial coin images from numismatic web sites.
As we are dealing with the problem of recognizing given the coin images, we did not consider the coins that are severely damaged or hard to recognize.
In the next step, we removed the background of each coin image by a standard background removal method and resized it to $256 \times 256$.
Each coin in the dataset has both observe (front) and reverse images.
For the purpose of the classification, we label the coin images according to their RIC \cite{Cohen1880-1892, Mattingly1923}.
We found that the coins with similar descriptions look similar to each other, making them almost impossible to differentiate.
Therefore, if the number of the different words in the descriptions for the two coins was less than a threshold (we set it to 2 in this paper), we considered them as the same class and assigned the same label.
Finally, we create a new coin dataset consisting of 4526 coins with RIC 314 labels and 96 Roman emperors.

{\bf Baseline method}:
As a baseline method, we use the SVM model as described in \cite{Kim2014}.
In \cite{Kim2014}, they extracted the SIFT descriptors \cite{Lowe2004} in the dense manner and used the $k$-means clustering method to build the visual code book.
Then, the image is represented as a histogram of the visual words from the codebook.
We also use the spatial pyramid model \cite{Lazebnik2006} to exploit the spatial information.
In this paper, we use the polar coordinate system as the spatial pyramid as it has shown the best performance in the previous ancient coin recognition approaches~\cite{Kim2014, Anwar2013}.
The polar coordinate system models $r$ radial scales and $\theta$ angular orientations.
We empirically use $r=2$ and $\theta = 6$.

{\bf Evaluation measure for classification}:
For measuring classification performance, we use 5-fold cross-validation with class-balanced partitions: we repeat the experiments 5 times with 4 subgroups as training data and 1 subgroup as test data so that each of 5 subgroups becomes the test data.
The classification accuracy is measure by the mean of the diagonal of the confusion matrix and we report the average of the 5 accuracies for 5 data splits.

{\bf CNN settings}:
We use the open-source package Caffe \cite{Jia2014} to implement our CNN model.
We follow the same network architecture as in \cite{Krizhevsky2012} except the final output layer where we replace the original 1000-way classification layer for our classification purpose (314-way for the RIC label prediction and 96-way for the Roman emperor prediction).
We also decrease the overall learning rate while increasing the learning rate for the new layer so that the rest of the model changes slowly while keeping a stronger pace of updates in the final layer \cite{Jia2014}.

\subsection{Coin Classification Results}

\begin{table}[t]
	\begin{center}
		\caption{Classification Accuracies for SVM and CNN}
		\label{tab:cnn_vs_svm}
		\begin{tabular}{|c|l|l|}
			\hline
			                   & SVM                  & CNN                        \\ \hline\hline
			 \texttt{Reverse}  & 42.76\% ($\pm 1.30$) & {\bf 62.28} \%($\pm 1.40$) \\ \hline
			 \texttt{Observe}  & 62.83\% ($\pm 2.42$) & {\bf 69.99} \%($\pm 2.53$) \\ \hline
			\texttt{Hierarchy} & 60.68\% ($\pm 1.23$) & {\bf 76.18} \%($\pm 2.01$) \\ \hline
		\end{tabular}
	\end{center}
\end{table}

We first discuss the fine-tuned CNN models that we use in this paper.
Figure \ref{fig:cnn_iter} depicts how the classification accuracy changes as a function of the iteration number (epoch).
As shown in the figure, the classification accuracy remains steady. after 40,000 iterations.
Therefore, we fine-tuned our CNN models over 50,000 iterations and use them thereafter.

The classification accuracy of the CNN model on the collected coin dataset is given in Table \ref{tab:cnn_vs_svm}.
\texttt{Reverse} presents the task to predict the RIC label given the reverse image.
In \texttt{Hierarchy}, we predict the RIC label given both the observe and reverse images using the hierarchical classification method as we discussed in Section \ref{s:pro_hi}.
We also show the classification accuracy for \texttt{Observe} which represents the task to predict the Roman emperor given the observe image.
Because the number of the emperors is less than the number of the RIC labels, \texttt{Observe} is easier than the other tasks.
\texttt{Hierarchy} can get benefit from performing the easy task (the emperor prediction) first followed by the more difficult task of RIC prediction.

CNN significantly outperforms SVM in all three tasks leading to up to 20\% increase in accuracy.
Specifically, CNN shows most significant improvement on \texttt{Reverse} side RIC classification for two reasons.  
First, there are significantly fewer emperors (96) than RIC labels (314).  
Next, the structure of \texttt{Reverse} side is typically more complex than that depicted on \texttt{Observe}, consisting of well-structured face profiles.
The convolutional feature of CNN is able to more effectively exploit the spatial information than the spatial histogram used in SVM.
Coins with the same RIC label have few consistent characteristic landmark regions and the CNN model is able to locate them effectively.
On the other hand, SVM has to depend on the fixed structure of the spatial pyramid model which may not be appropriate for some specific RIC labels.
We will discuss the recovery of the discriminative regions found by the CNN models in Section \ref{s:dr01}.

\begin{figure}
	\begin{center}
		\begin{subfigure}[b]{0.24\linewidth}
			\includegraphics[width=\textwidth]{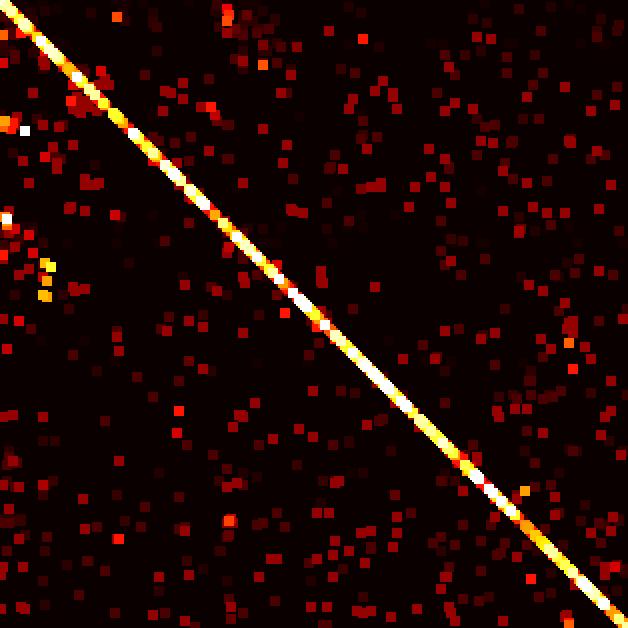}
			\caption{CNN \\~~~\texttt{Reverse}}
		\end{subfigure}~
		\begin{subfigure}[b]{0.24\linewidth}
			\includegraphics[width=\textwidth]{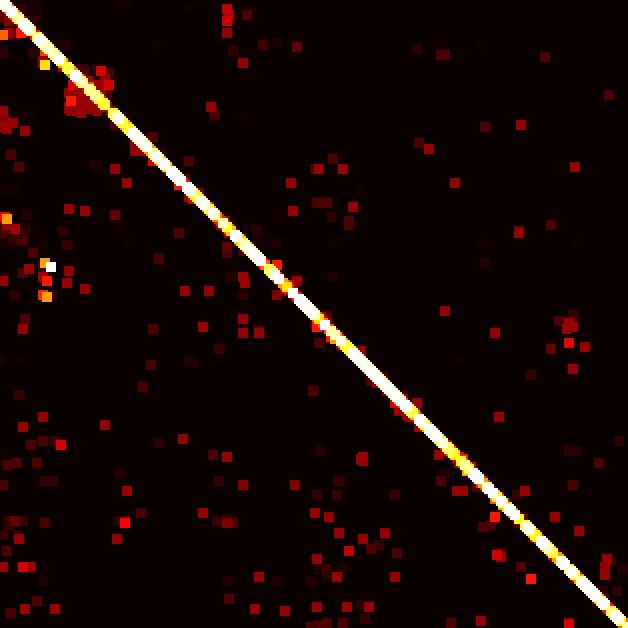}
			\caption{CNN \\~~~\texttt{Hierarchy}}
		\end{subfigure}~
		\begin{subfigure}[b]{0.24\linewidth}
			\includegraphics[width=\textwidth]{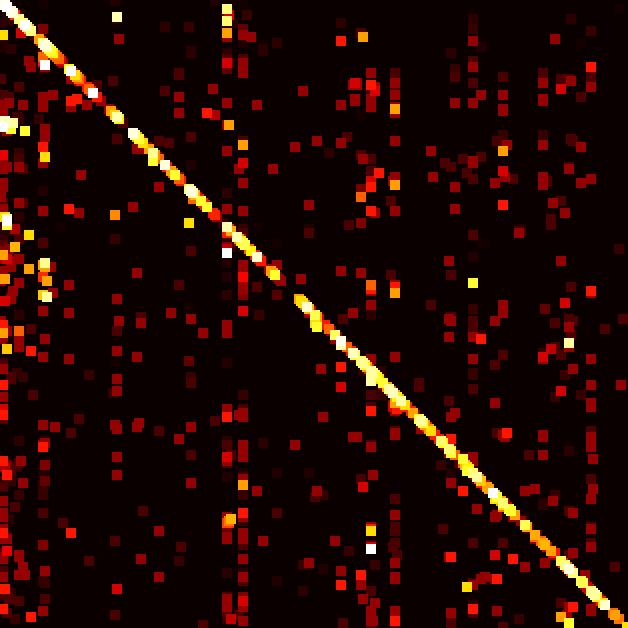}
			\caption{SVM \\~~~\texttt{Reverse}}
		\end{subfigure}~
		\begin{subfigure}[b]{0.24\linewidth}
			\includegraphics[width=\textwidth]{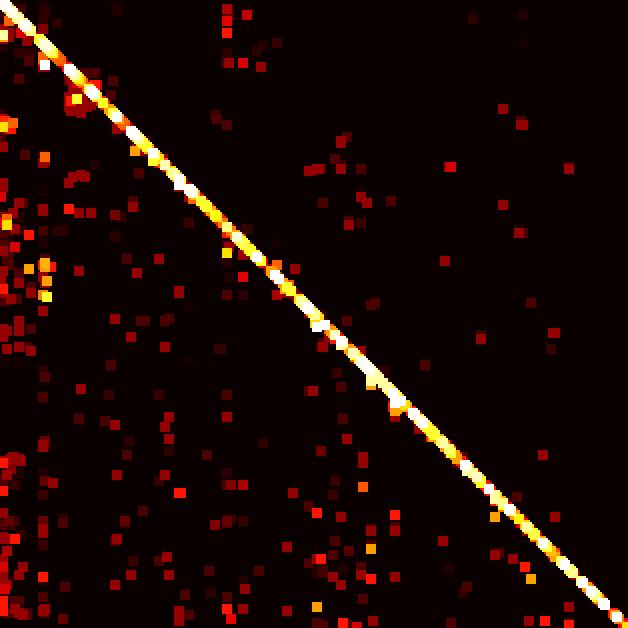}
			\caption{SVM \\~~~\texttt{Hierarchy}}
		\end{subfigure}
	\end{center}
	\caption{Confusion matrices of CNN and SVM for \texttt{Reverse} and \texttt{Hierarchy}. In both models, \texttt{Hierarchy} performs better than \texttt{Reverse}. CNN \texttt{Hierarchy} has improved the classification accuracies across all the RIC labels as it takes an advantage of the hierarchical structure of the RIC labels. For visualization, the smoothed heat map is used.}
	\label{fig:confumat}
\end{figure}

\begin{figure}[t]
	\centering
	\includegraphics[width=0.8\linewidth]{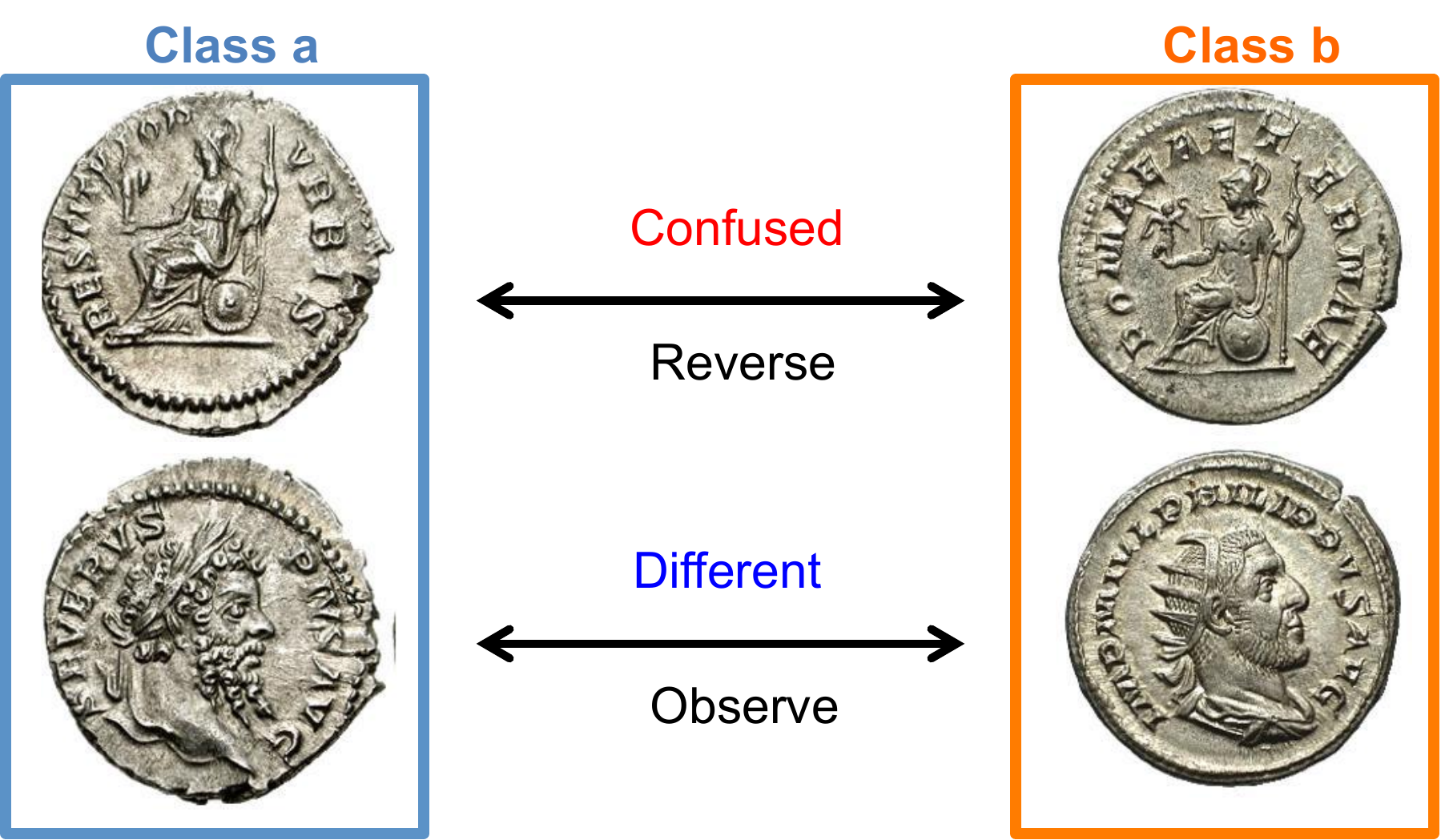}
	\caption{Example of two classes where confusion caused by \texttt{Reverse} is resolved by \texttt{Hierarchy}. The reverse images for the two classes look similar to each other. On the contrary, the observe images make them distinguishable.}
	\label{fig:both_improves}
\end{figure}

The confusion matrices for the classification of the RIC label are depicted in Figure \ref{fig:confumat}.
\texttt{Hierarchy} outperforms \texttt{Reverse} in both CNN and SVM models as it exploits the hierarchical structure of the RIC labels.
To better understand this phenomenon, we select two classes that are confused by \texttt{Reverse} but \texttt{Hierarchy} can distinguish them correctly as shown in Figure \ref{fig:both_improves}.
The confusion caused by the similarity between the reverse images can be removed using the differently depicted observe images.

\subsection{Discriminative Regions and Landmarks \label{s:dr01}}

\begin{figure*}
	\begin{center}
		\begin{tabular}{cc@{~}c@{~}c@{~}c@{~}c}
			& $\epsilon=0.1$& $\epsilon=0.3$& $\epsilon=0.5$& $\epsilon=0.7$&$\epsilon=1$\\
			\includegraphics[scale=0.3]{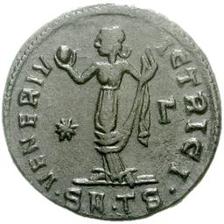} & 
			\includegraphics[scale=0.3]{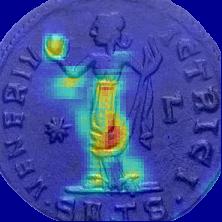} & 
			\includegraphics[scale=0.3]{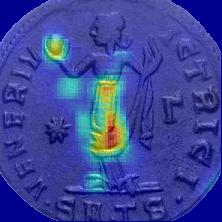} & 
			\includegraphics[scale=0.3]{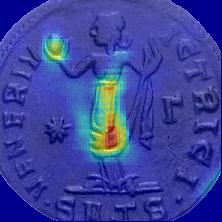} & 
			\includegraphics[scale=0.3]{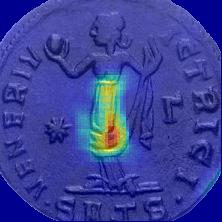} & 
			\includegraphics[scale=0.3]{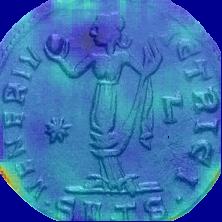}\\
			& 
			{\scriptsize $p_c(f_\mathbf{\mathbf{I}}(\mathbf{x}^*)) = 0.907$} & 
			{\scriptsize $p_c(f_\mathbf{\mathbf{I}}(\mathbf{x}^*)) = 0.734$} & 
			{\scriptsize $p_c(f_\mathbf{\mathbf{I}}(\mathbf{x}^*)) = 0.616$} & 
			{\scriptsize $p_c(f_\mathbf{\mathbf{I}}(\mathbf{x}^*)) = 0.330$} & 
			{\scriptsize $p_c(f_\mathbf{\mathbf{I}}(\mathbf{x}^*)) = 0.002$}\\
			&&&&&\\
			& $\epsilon=0.1$& $\epsilon=0.3$& $\epsilon=0.5$& $\epsilon=0.7$&$\epsilon=1$\\
			\includegraphics[scale=0.3]{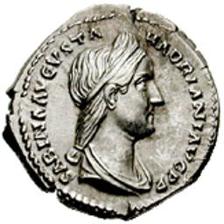} & 
			\includegraphics[scale=0.3]{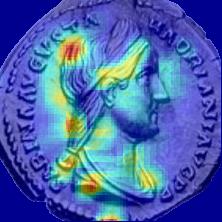} & 
			\includegraphics[scale=0.3]{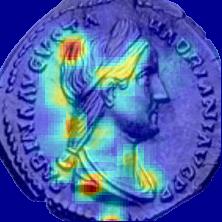} & 
			\includegraphics[scale=0.3]{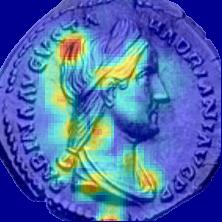} & 
			\includegraphics[scale=0.3]{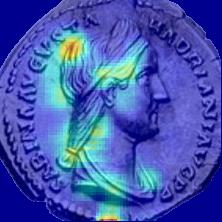} & 
			\includegraphics[scale=0.3]{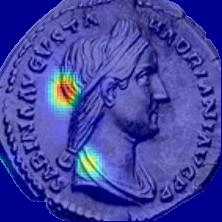}\\
			& 
			{\scriptsize $p_c(f_\mathbf{\mathbf{I}}(\mathbf{x}^*)) = 0.912$} & 
			{\scriptsize $p_c(f_\mathbf{\mathbf{I}}(\mathbf{x}^*)) = 0.791$} & 
			{\scriptsize $p_c(f_\mathbf{\mathbf{I}}(\mathbf{x}^*)) = 0.712$} & 
			{\scriptsize $p_c(f_\mathbf{\mathbf{I}}(\mathbf{x}^*)) = 0.561$} & 
			{\scriptsize $p_c(f_\mathbf{\mathbf{I}}(\mathbf{x}^*)) = 0.000$}\\
		\end{tabular}
		\caption{Visualizatoin of $\mathbf{x}^*$ as a function of $\epsilon$, where $\mathbf{x}^*$ is the optimized solution. As $\epsilon$ increase, classification probability $p_c(\cdot)$  for class $c$ and landmark areas decrease. All the masked images are correctly identified by the CNNs except $\epsilon = 1$ because it places no constraint for the correct classification and may lead to wrong prediction results. For visualization, we rescale $\mathbf{x}^*$ to the range of $[0, 1]$.}
		\label{fig:var_eps_both}
	\end{center}
\end{figure*}

In this section, we first examine how the selected regions and confidence values from the CNN model change as a function of $\epsilon$.
For this purpose, we choose one reverse and one observe images and vary $\epsilon$ from $0.1$ to $1.0$.
Note that $\epsilon = 1$ implies that the constraint in (\ref{eqn:opt02}) will never be violated.

Figure \ref{fig:var_eps_both} shows the visualization of discovered landmarks as a function of $\epsilon$.
Because larger $\epsilon$ allows smaller confidence value, the total area of the characteristic parts becomes smaller, {\it i.e.} very essential parts are remained.
Therefore as we increase $\epsilon$, relatively less significant regions are first removed on the coin.
For example, Venus in the upper panel holds a small apple which is considered as the characteristic part at first.
However, as we increase $\epsilon$, the size of the discriminative areas becomes smaller and finally the apple turns out less significant than the toss.

When $\epsilon = 1$, no constraint is placed during the optimization.
Therefore, the gradient decent method tries to find the mask as sparse as possible without considering the correct prediction, having the discovered regions meaningless.

On the other hand, the discriminative regions on the observe images change slowly.
Unlike the reverse where different characteristic symbols appear in variable locations, the observe images have common structures, {\it i.e.} profiles of the Roman emperors.
Therefore, the observe images need more parts to remain distinguishable from the others than the reverse images.
As shown in Figure \ref{fig:var_eps_both}, head and bust remains present for all $\epsilon$ values.

\begin{figure*}[t]
	\begin{center}
		\begin{tabular}{c@{~}>{\centering\arraybackslash}m{52pt}|>{\centering\arraybackslash}m{52pt}@{~}>{\centering\arraybackslash}m{52pt}|>{\centering\arraybackslash}m{52pt}@{~}>{\centering\arraybackslash}m{52pt}@{~}|>{\centering\arraybackslash}m{52pt}@{~}>{\centering\arraybackslash}m{52pt}|l}
			&& \multicolumn{2}{c|}{Proposed method} & \multicolumn{2}{c|}{\cite{Zhou2015}} & \multicolumn{2}{c|}{\cite{Simonyan2014}} & Description\\
			\hline
			& &$11 \times 11$    & $21 \times 21$    & $11 \times 11$    & $21 \times 21$    & $11 \times 11$    & $21 \times 21$  \\
			
			(a) &
			\includegraphics[scale=0.23]{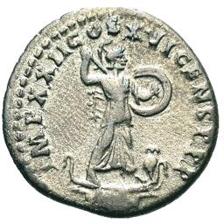} & 
			\includegraphics[scale=0.23]{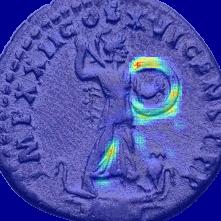} & 
			\includegraphics[scale=0.23]{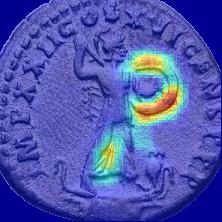} & 
			\includegraphics[scale=0.23]{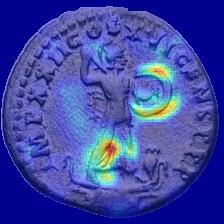} & 
			\includegraphics[scale=0.23]{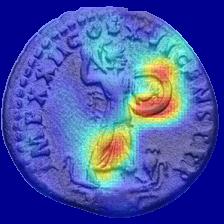} & 
			\includegraphics[scale=0.23]{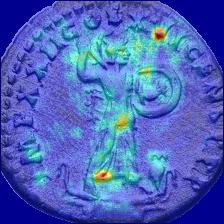} & 
			\includegraphics[scale=0.23]{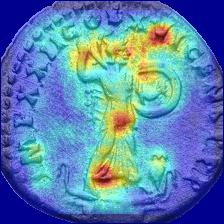} & 
			{\scriptsize \pbox{15em}{standing\\capital\\column\\spear\\shield\\owl}}\\ 
			(b) &
			\includegraphics[scale=0.23]{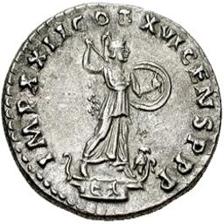} & 
			\includegraphics[scale=0.23]{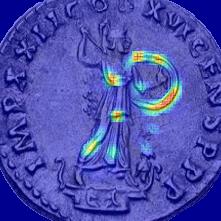} & 
			\includegraphics[scale=0.23]{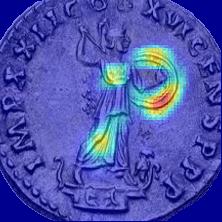} & 
			\includegraphics[scale=0.23]{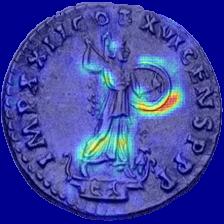} & 
			\includegraphics[scale=0.23]{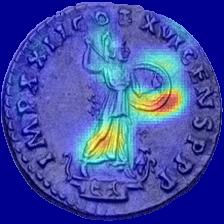} & 
			\includegraphics[scale=0.23]{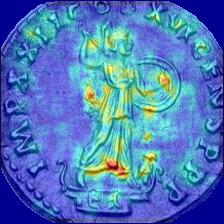} & 
			\includegraphics[scale=0.23]{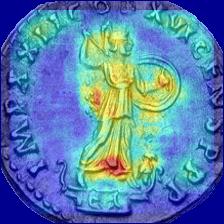} & 
			{\scriptsize \pbox{15em}{standing\\capital\\column\\spear\\shield\\owl}}\\ 
			(c) &
			\includegraphics[scale=0.23]{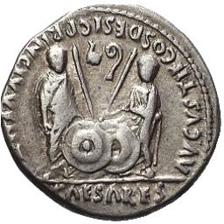} & 
			\includegraphics[scale=0.23]{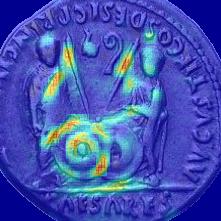} & 
			\includegraphics[scale=0.23]{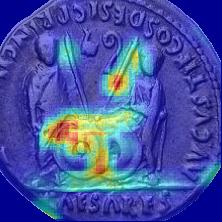} & 
			\includegraphics[scale=0.23]{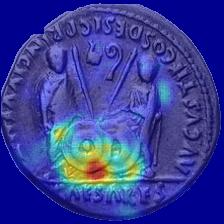} & 
			\includegraphics[scale=0.23]{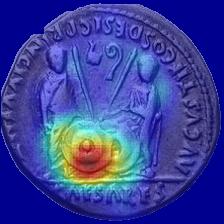} & 
			\includegraphics[scale=0.23]{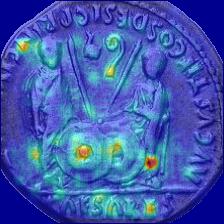} & 
			\includegraphics[scale=0.23]{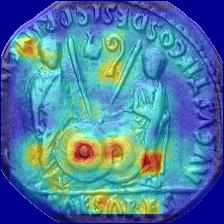} & 
			{\scriptsize \pbox{15em}{stand\\hand\\shield\\spear\\simpulum\\lituus}}\\ 
			(d) &
			\includegraphics[scale=0.23]{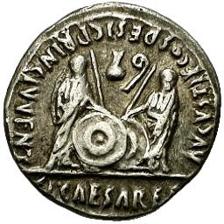} & 
			\includegraphics[scale=0.23]{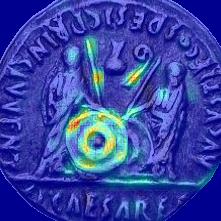} & 
			\includegraphics[scale=0.23]{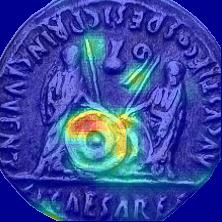} & 
			\includegraphics[scale=0.23]{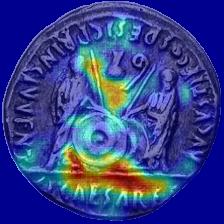} & 
			\includegraphics[scale=0.23]{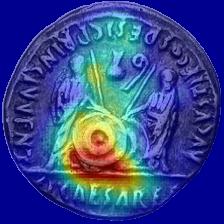} & 
			\includegraphics[scale=0.23]{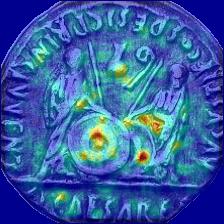} & 
			\includegraphics[scale=0.23]{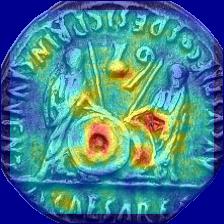} & 
			{\scriptsize \pbox{15em}{stand\\hand\\shield\\spear\\simpulum\\lituus}}\\ 
			(e) &
			\includegraphics[scale=0.23]{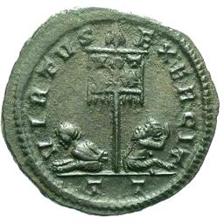} & 
			\includegraphics[scale=0.23]{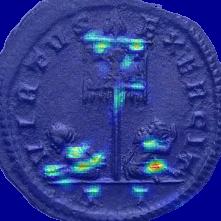} & 
			\includegraphics[scale=0.23]{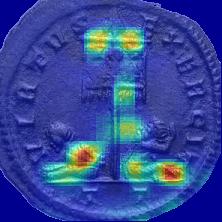} & 
			\includegraphics[scale=0.23]{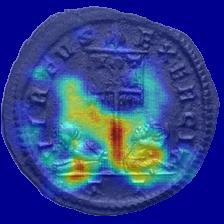} & 
			\includegraphics[scale=0.23]{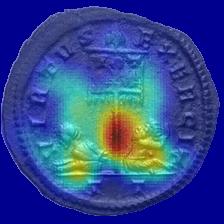} & 
			\includegraphics[scale=0.23]{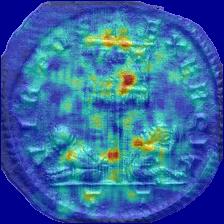} & 
			\includegraphics[scale=0.23]{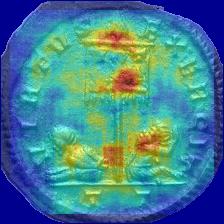} & 
			{\footnotesize \pbox{15em}{drapery\\captives}}\\ 
			(f) &
			\includegraphics[scale=0.23]{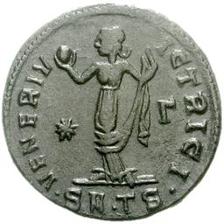} & 
			\includegraphics[scale=0.23]{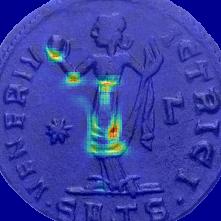} & 
			\includegraphics[scale=0.23]{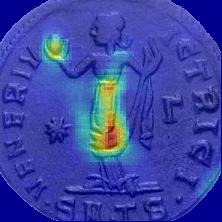} & 
			\includegraphics[scale=0.23]{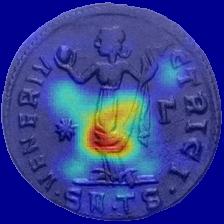} & 
			\includegraphics[scale=0.23]{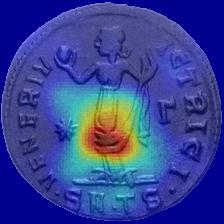} & 
			\includegraphics[scale=0.23]{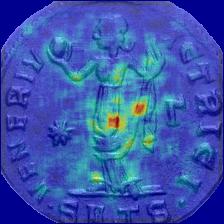} & 
			\includegraphics[scale=0.23]{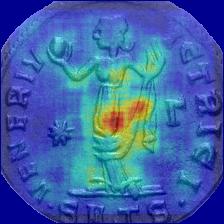} & 
			{\footnotesize \pbox{15em}{standing\\head\\hand\\apple\\drapery}}\\

			(g) &
			\includegraphics[scale=0.23]{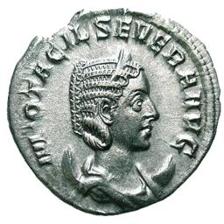} & 
			\includegraphics[scale=0.23]{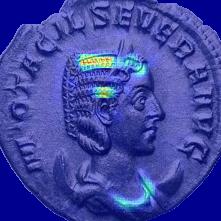} & 
			\includegraphics[scale=0.23]{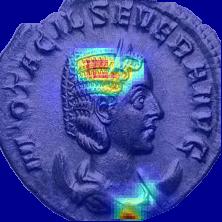} & 
			\includegraphics[scale=0.23]{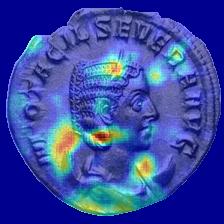} & 
			\includegraphics[scale=0.23]{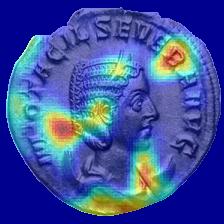} & 
			\includegraphics[scale=0.23]{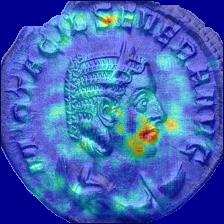} & 
			\includegraphics[scale=0.23]{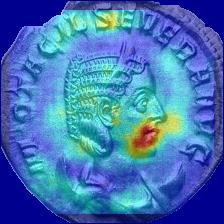} & 
			{\footnotesize \pbox{15em}{draped\\bust\\crescent}}\\ 
			(h) &
			\includegraphics[scale=0.23]{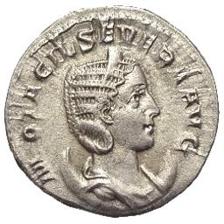} & 
			\includegraphics[scale=0.23]{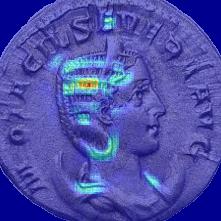} & 
			\includegraphics[scale=0.23]{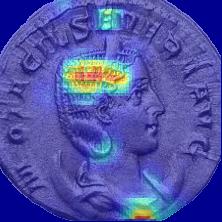} & 
			\includegraphics[scale=0.23]{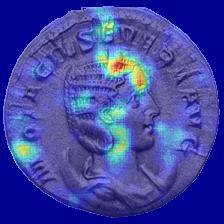} & 
			\includegraphics[scale=0.23]{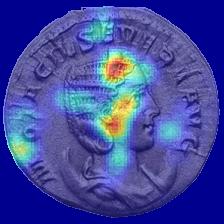} & 
			\includegraphics[scale=0.23]{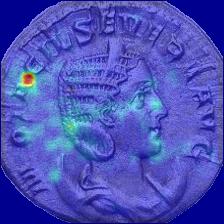} & 
			\includegraphics[scale=0.23]{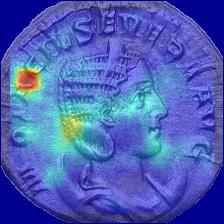} & 
			{\footnotesize \pbox{15em}{draped\\bust\\crescent}}\\ 
			(i) &
			\includegraphics[scale=0.23]{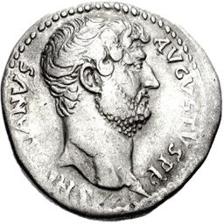} & 
			\includegraphics[scale=0.23]{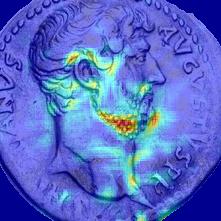} & 
			\includegraphics[scale=0.23]{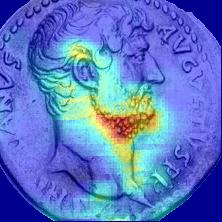} & 
			\includegraphics[scale=0.23]{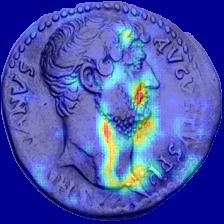} & 
			\includegraphics[scale=0.23]{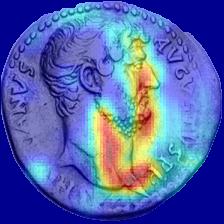} & 
			\includegraphics[scale=0.23]{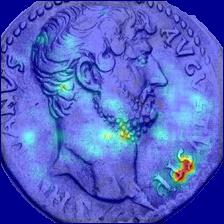} & 
			\includegraphics[scale=0.23]{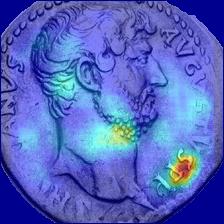} & 
			{\footnotesize \pbox{15em}{bare\\head}}\\ 
			(j) &
			\includegraphics[scale=0.23]{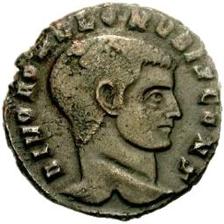} & 
			\includegraphics[scale=0.23]{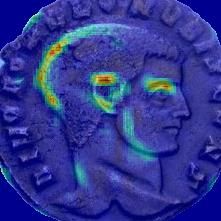} & 
			\includegraphics[scale=0.23]{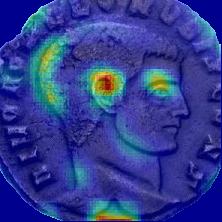} & 
			\includegraphics[scale=0.23]{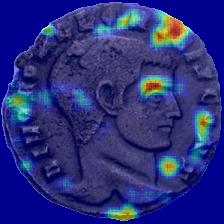} & 
			\includegraphics[scale=0.23]{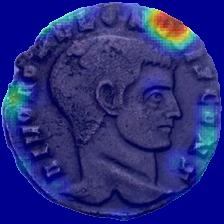} & 
			\includegraphics[scale=0.23]{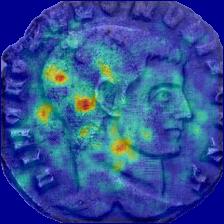} & 
			\includegraphics[scale=0.23]{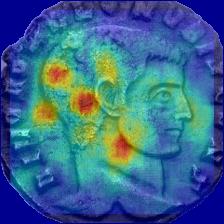} & 
			{\footnotesize \pbox{15em}{bare\\head}}\\ 
		\end{tabular}
		\caption{Visualization of discovered landmarks for reverse and observe images. Red denotes more discriminative, blue less significant. The proposed method and \cite{Zhou2015} agree with each other while the saliency extraction~\cite{Simonyan2014} focuses on strong edge areas. Note that the discovered regions are correlated with descriptions from human expert annotations. For visualization, we rescale $\mathbf{x}^*$ to the range of $[0, 1]$.}
		\label{fig:rev_var01}
	\end{center}
\end{figure*}

Figure \ref{fig:rev_var01} depicts the visualization of the discovered landmarks on both reverse and observe images with two different sliding windows ($11\times11$ and $21\times21$).
We set $\epsilon$ to 0.5 and choose the coins that are correctly classified by the CNNs for the experiments.
The results confirm that the proposed method is robust with respect to the window sizes.
Moreover, the coins with the same RIC label, (a) and (b), (c) and (d), (g) and (h) in Figure \ref{fig:rev_var01}, share the similar landmarks.
The results imply that there exists a set of characteristic regions per class, class-specific discriminative regions.   
As we will see next, such regions indeed point to intuitive visual landmarks associated with RIC descriptions.

{\bf Qualitative analysis}: 
There is no ground truth information available for the discriminative regions.
Therefore, we qualitatively analyze our proposed method with two different schemes.
First, we qualitatively compare our proposed method with recently proposed approaches \cite{Zhou2015, Simonyan2014}.
In \cite{Zhou2015}, they identify which regions of the image lead to the high unit activations by replicating an image many times with small occluders at different locations in the image and measuring the discrepancy of the unit activations between the original image and the occluded images.
An image patch which leads to large discrepancy can be considered as important to the unit.
We use two different sizes of the image patches ($11\times11$ and $21\times21$) and measure the difference of the class score ($S_c$ in (\ref{eqn:loss_wrt_x1})) on a dense grid with a stride of 3.

The saliency extraction method which computes the gradient of the CNNs with respect to the image was proposed in \cite{Simonyan2014}.
The single pass of the back propagation is used to find the saliency map.
For fair comparison, we perform the moving average with the same patch size as the other experiments subsequent to back propagation.

The experimental results in Figure \ref{fig:rev_var01} show that our method and \cite{Zhou2015} largely agree with each other.
This implies that our method is able to find the important regions which lead to large discrepancy or, equivalently, significant changes in classification accuracy.
On the other hand, the saliency extraction method~\cite{Simonyan2014} tries to find strong edge areas without considering class-characteristics.
For example, it fails to find the shield in Figure \ref{fig:rev_var01}a and \ref{fig:rev_var01}b, which both our method and \cite{Zhou2015} are able to discover.

Nevertheless, our proposed method has a distinct advantage over the occlusion-based approach as in \cite{Zhou2015} in terms of computational time.
The method in \cite{Zhou2015} requires a very large number of CNN evaluations ({e.g.}, more than 5000 for an image of $256 \times 256$).
On the other hand, the proposed method is based on the optimization formulation, usually converging in fewer than 100 iterations, while identifying qualitatively similar landmarks.

Next, we use the coin descriptions from RIC to analyze the proposed method.
For this purpose, we remove stop words in the RIC descriptions and list the remaining words as depicted in Figure \ref{fig:rev_var01}.
The selected landmarks by the proposed method strongly correlate with the descriptions, such as the shield found in Figure \ref{fig:rev_var01}a, \ref{fig:rev_var01}b, \ref{fig:rev_var01}c and \ref{fig:rev_var01}d.
On the other hand, the apple in Figure \ref{fig:rev_var01}f is successfully found by our method while the others fail to find it or discover it with little confidence.
This attests to the practical utility of the proposed approach in identifying the landmarks consistent with human expert annotations.
In addition, the proposed method may assist non-experts in generating a visual guidebook to identify the ancient Roman coins without specific domain expertise.

\section{Conclusion \label{s:con}}
We proposed a novel method to discover the characteristic landmarks of the ancient Roman imperial coins.
Our method automatically finds the smallest set of the discriminative regions sufficient to represent the identity of the full image and distinguish it from other available classes.
The qualitative analysis on the visualization of the discovered regions confirm that the proposed method is able to effectively find the class-specific regions but also it is consistent with the human expert annotations.
The proposed framework to identify the ancient Roman imperial coins outperforms the previous approach in the domain of the coin classification by using the hierarchical structure of the RIC labels.

\clearpage

{\small
\bibliographystyle{ieee}
\bibliography{egbib}
}

\end{document}